\documentclass[11pt]{article}

\usepackage[preprint]{acl}

\usepackage{url}
\usepackage{times}
\usepackage{latexsym}

\usepackage[T1]{fontenc}

\usepackage[utf8]{inputenc}

\usepackage{microtype}

\usepackage{inconsolata}

\usepackage{graphicx}

\usepackage{tikz}
\usepackage{pgfplots}
\usepgfplotslibrary{polar}
\pgfplotsset{compat=1.18}
\usepackage[table]{xcolor}
\usepackage{xcolor}
\usepackage{caption}
\usepackage{pgf-pie}

\usepackage{booktabs}
\usepackage{multirow}

\usepackage{placeins}
\usepackage{longtable}

\title{\textsc{LuxMT} Technical Report}

\author{Nils Rehlinger \\
  University of Luxembourg / Esch-Belval, Esch-sur-Alzette, Luxembourg \\
  \texttt{nils.rehlinger@uni.lu} 
  }
\begin{document}
\maketitle
\begin{abstract}
We introduce \textsc{LuxMT}, a machine translation system based on \textsc{Gemma 3} 27B and fine-tuned for translation from Luxembourgish (LB) into French (FR) and English (EN). To assess translation performance, we construct a novel benchmark covering LB-FR, LB-EN, and LB-FR using human-translated data from Luci, a tourist magazine about Luxembourg. Training data stems from \textsc{LuxAlign}, a parallel corpus of multilingual Luxembourgish news articles, and LB parliamentary transcripts augmented with \textsc{Google Translate}. We filter the data using \textsc{LuxEmbedder}, LB sentence embeddings, to remove low-equivalence segment-pairs. Overall, LuxMT's results suggest strong improvements over the \textsc{Gemma 3} baseline, even for translating LB to German (DE), despite the training data not containing any DE. We also explore \textsc{LuxEmbedder}'s potential to be used as a quality estimation metric and find strong correlations with other reference-based metrics. However, we call for further research to fully assess the metric's utility and advise using it with caution.
\end{abstract}

\section{Introduction}

We present \textsc{LuxMT}\footnote{\url{https://luxasr.uni.lu/luxmt}},
a fine-tune of \textsc{Gemma 3} specialised for machine translation (MT) from Luxembourgish (LB) into French (FR) and English (EN) \cite{team2025gemma}. This paper describes the model selection process, dataset curation, and fine-tuning process. In addition to the language pairs mentioned above, we also evaluate \textsc{LuxMT}'s performance on LB to German (DE) translation.

Prior to selecting a model, we construct a benchmark to compare out-of-the-box MT models and select the best performing base model for subsequent fine-tuning.

\section{Benchmark}
While \textsc{FLORES-200} is widely used to benchmark multilingual MT \cite{costa2022no}, we opt to develop a custom benchmark to mitigate the risk of data contamination. Base models are trained on large-scale web-scraped corpora and it is possible that \textsc{FLORES-200} was inadvertently included in the base models' training data. Such contamination would risk skewing the results \cite{xu2024benchmark}.

The data is based on Luci\footnote{\url{https://www.visitluxembourg.com/luci-order/}}, a quadrilingual (LB, FR, DE, EN) tourist magazine for Luxembourg, editions 1-6\footnote{Due to copyright restrictions, the benchmark is currently not publicly available.}.
We segment the text using spaCy\footnote{\url{https://www.spacy.io/}} and align segment pairs with \textsc{LuxEmbedder}, Luxembourgish sentence embeddings \cite{philippy2025luxembedder}, yielding three parallel corpora: LB-FR, LB-EN, and LB-DE. We then standardise quotation marks for each language, remove segment pairs with source segments shorter than five words, and remove duplicates. Finally, we sort the segment pairs from highest to lowest based on cosine similarity scores using \textsc{Luxembedder}, keep the top 500 segment pairs per language pair, and manually verify them. Some segment pairs required adjustment on a segment level as dislocating segments from their context can diminish the equivalence of segment pairs, e.g., usage of proper names in the source vs. pronouns in the translation.

\section{Evaluation Metrics}
As each metric has its own strengths and weaknesses, researchers recommend using multiple complementary metrics to obtain robust and reliable evaluation results \cite{kocmi2024navigating, moghe2025machine}. Accordingly, we construct an ensemble\footnote{\url{https://github.com/greenirvavril/lux-eval}}
by choosing neural metrics each trained on different types of human evaluation data (\textsc{BLEURT-20}, \textsc{xCOMET\textsuperscript{XL}}; \citealt{pu2021learning, guerreiro2024xcomet}), a metric based on cosine similarity (\textsc{BERTScore}; \citealt{zhang2019bertscore}), an \textit{n}-gram surface overlap metric (\textsc{BLEU}; \citealt{papineni2002bleu}), a character-level surface-overlap metric (\textsc{chrF2}; \citealt{popovic2015chrf, popovic2016chrf}), and translation edit-rate (\textsc{TER}; \citealt{snover2006study}). Since surface-overlap metrics are found to be questionable for inter-system comparison, we exclude them from the main report and only report these scores in the appendix \cite{kocmi2024navigating}.

Since LB is a low-resource language in MT, there are currently no metrics trained for LB. Consequently, we are limited to reference-based metrics. Nevertheless, \textsc{LuxEmbedder}'s cosine similarity score has the potential to be used as a reference-less metric, aka. quality estimation (QE). To investigate its utility for MT evaluation, we include \textsc{LuxEmbedder} as an experimental QE metric. We normalise the \textsc{LuxEmbedder} scores between 80 and 100 to scale with the other metrics.

To aid score interpretation, we use \citet{kocmi2024navigating}'s \textsc{MT Thresholds}, which maps and converts metric score differences to estimated human pairwise accuracy. This conversion enables a more direct comparison between scores. For example, a difference of one \textsc{BLEURT-20} point has an estimated accuracy of 78.5\% with humans, meaning that there is a 78.5\% probability that a human would be able to tell which system has the higher \textsc{BLEURT-20} score. In turn, this equals a 0.58 difference in \textsc{BERTScore} points. We do not report accuracy estimates for \textsc{TER} and \textsc{LuxEmbedder} as the conversion tool does not support these metrics.

\section{Analysis}
We compare system-level performances and determine statistical significance using paired bootstrap resampling with a 95\% confidence interval computed for the difference in average metric scores between each candidate system and the baseline \cite{koehn2004statistical}.

\section{Model Selection}
The goal of the model selection process is to identify the best-performing local model for further fine-tuning. We compare various popular local large language models (LLM) by evaluating them on the Luci benchmark for LB$\rightarrow$EN, LB$\rightarrow$FR, and LB$\rightarrow$DE. The models included are \textsc{Gemma 3} \cite{team2025gemma}, \textsc{Aya Expanse} \cite{dang2024aya}, \textsc{Command R}\footnote{\url{https://docs.cohere.com/docs/command-r}}, \textsc{Llama 3.1} \cite{grattafiori2024llama}, \textsc{Llama 4}\footnote{\url{https://ai.meta.com/blog/llama-4-multimodal-intelligence/}}, \textsc{Mistral Small 3.2}\footnote{\url{https://huggingface.co/mistralai/Mistral-Small-3.2-24B-Instruct-2506}}, and \textsc{Phi 4} \cite{abdin2024phi}.
Each model receives the same prompt:

'Translate from Luxembourgish to [TARGET LANGUAGE]. Only provide the translation, nothing else: [source\_segment]'.

All metrics unanimously indicate that \textsc{Gemma 3} is the best performing model (see Table~\ref{tab:base_models_eval} for average system scores and Appendix~\ref{sec:full-results} for the full results).

\begin{table}[t]
\small
\begin{tabular}{lcccc}
\toprule
System & \textsc{LE}$\uparrow$ & \textsc{BS}$\uparrow$ & \textsc{B-20}$\uparrow$ & \textsc{xC\textsuperscript{XL}}$\uparrow$ \\
\midrule
\textsc{Gemma 3}        & \textbf{93.9} & \textbf{93.6} & \textbf{74.0} &  \textbf{89.7} \\
\rowcolor{gray!20}
\textsc{Aya Expanse}    & 90.0 & 91.7 & 67.0 & 82.2  \\
\textsc{Command R}      & 92.5 & 91.3 & 66.4 & 81.6 \\
\rowcolor{gray!20}
\textsc{Llama 3.1}      & 93.0 & 93.0 & 71.0 & 87.3  \\
\textsc{Llama 4}        & 93.5 & 92.9 & 70.9 & 86.7  \\
\rowcolor{gray!20}
\textsc{Mistral S}      & 90.4 & 91.6 & 65.8 & 80.4  \\
\textsc{Phi 4}          & 85.6 & 90.2 & 61.2 & 73.5 \\
\bottomrule
\end{tabular}
\caption{Automatic evaluation results assessing various popular out-of-the-box local LLMs on the Luci benchmark for LB$\rightarrow$EN ($N=500$), LB$\rightarrow$FR ($N=500$), and LB$\rightarrow$DE ($N=500$). The results are averaged across the three language pairs. Metrics include \textsc{LuxEmbedder} (LE), \textsc{BERTScore} (BS), \textsc{BLEURT-20} (B-20), and \textsc{xCOMET\textsuperscript{XL}} (xC\textsuperscript{XL}). $\uparrow$ indicates that higher scores are better. The highest scores are highlighted in \textbf{bold}.}
\label{tab:base_models_eval}
\end{table}

\section{Fine-tuning}
\subsection{Data}
Our data sources are \textsc{LuxAlign} \cite{philippy2025luxembedder} and transcripts of Luxembourgish parliamentary debates provided by the Chambre des Députés. \textsc{LuxAlign} \cite{philippy2025luxembedder} is a segment-pair-level aligned parallel corpus of RTL\footnote{Radio Television Luxembourg is the largest news outlet in Luxembourg. \url{https://rtl.lu/}} articles for language pairs LB-FR and LB-EN. We remove duplicates and segment pairs with less than five words in the source segment. Lastly, we augment the monolingual parliament data by translating it into EN and FR using \textsc{Google Translate}\footnote{Data augmented in August 2025.}.

\subsection{Preliminary Testing}
\subsubsection{Filtering}
To select filtering methods for data curation and hyperparameters for fine-tuning, we conduct a series of exploratory tests. As evaluation metrics, we use \textsc{BERTScore} \textsc{BLEURT-20}, \textsc{BLEU}, \textsc{chrF2}, and \textsc{TER}.\footnote{At this stage, the benchmark was limited to 400 segment pairs from Luci for LB$\rightarrow$FR. and 233 segment pairs for LB$\rightarrow$EN. We also used a more limited evaluation metric ensemble.}
For training data curation, we experiment with fine-tuning \textsc{Gemma 3} on varying \textsc{LuxEmbedder} thresholds to filter \textsc{LuxAlign}: .90, .95, and .99 (see Table~\ref{tab:automatic-eval_thresholds}).
We use a learning rate of 2e-5 and fine-tune the model for one epoch. Even though a threshold of .99 drastically reduces the size of the \textsc{LuxAlign} corpus by 85.2\%,
the results suggest that it leads to the best performance. This may be due to RTL articles not being 1-to-1 translations and most of the corpus therefore not being suitable for fine-tuning for MT purposes.

\begin{table}[t]
\centering
\small
\begin{tabular}{lcc}
\toprule
System & \textsc{BS}$\uparrow$ & \textsc{B-20}$\uparrow$ \\
\midrule
\textsc{Gemma 3} & 93.8 & 77.2 \\
\rowcolor{gray!20}
\textsc{LE} .90      & 93.6 & 77.6 \\
\textsc{LE} .95      & 94.0 & 78.1 \\
\rowcolor{gray!20}
\textsc{LE} .99      & \textbf{94.4} & \textbf{78.8} \\
\bottomrule
\end{tabular}
\caption{Automatic evaluation results assessing various \textsc{LuxEmbedder} (LE) filter thresholds, using \textsc{BERTScore} (BS) and \textsc{BLEURT-20} (B-20). Results are based on average system scores for LB$\rightarrow$FR ($N=400$) and LB$\rightarrow$EN ($N=233$) on Luci segment pairs.}
\label{tab:automatic-eval_thresholds}
\end{table}

\subsubsection{Temperature}
We also explore temperatures 0.1-1.0.
We find that the results are inconclusive and decide to use \textsc{Gemma 3}'s standard temperature setting of 1.0 (see Appendix~\ref{sec:appendix_temp}).

\subsubsection{Fine-tuning Schedule}
We experiment with the number of epochs for fine-tuning. We fine-tune models for 1-3 epochs and compare their performances. The results are summarised in Table~\ref{tab:automatic-eval_epoch}.
We find that fine-tuning the model for one epoch yields the best results.

\begin{table}[t]
\centering
\small
\begin{tabular}{lcc}
\toprule
Epochs & \textsc{BS}$\uparrow$ & \textsc{B-20}$\uparrow$ \\
\midrule
1     & \textbf{94.4} & \textbf{78.8} \\
\rowcolor{gray!20}
2     & \textbf{94.4} & 78.5 \\
3     & 94.2 & 78.4 \\
\bottomrule
\end{tabular}
\caption{Automatic evaluation results assessing fine-tuning \textsc{Gemma 3} for 1-3 epochs. Results are based on average system scores for LB$\rightarrow$FR ($N=400$) and LB$\rightarrow$EN ($N=233$) on Luci segment pairs.}
\label{tab:automatic-eval_epoch}
\end{table}

\subsection{Final Model}

Before adding the augmented data to the final fine-tuning mixture, we filter it with a \textsc{LuxEmbedder} threshold of .98. The rational behind lowering the threshold in comparison to the threshold applied to \textsc{LuxAlign} is that the translations are 1-to-1 and a lower threshold preserves more data. The final fine-tuning data mixture is summarised in Figure~\ref{fig:data-mixture-pie}.
The prompt used for fine-tuning is:

Translate from Luxembourgish to [Target Language]: [source segment]

The model is fine-tuned for one epoch with a learning-rate of 2e-5.

\pgfplotsset{compat=1.18}

\definecolor{frRTL}{RGB}{147,211,215}  
\definecolor{frChD}{RGB}{43,186,215}   
\definecolor{enRTL}{RGB}{240,167,34}  
\definecolor{enChD}{RGB}{240,102,53}   

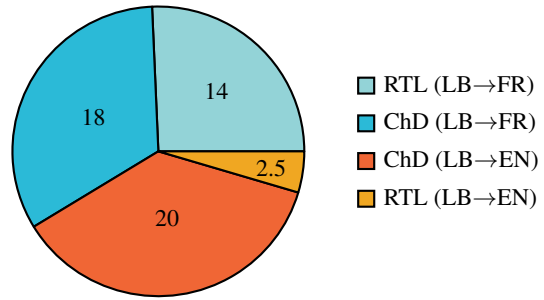
\begin{figure}[t]
    \centering
    \footnotesize
    \begin{tikzpicture}[scale=0.80]
        \pie[
            radius=2.4,
            text=legend,
            sum=auto,
            color={frRTL, frChD, enChD, enRTL}
        ]{
            14/RTL (LB$\rightarrow$FR),
            18/ChD (LB$\rightarrow$FR),
            20/ChD (LB$\rightarrow$EN),
            2.5/RTL (LB$\rightarrow$EN)
        }
    \end{tikzpicture}
    \caption{Data mixture for MT fine-tuning.
    LB$\rightarrow$FR total: 32k; LB$\rightarrow$EN total: 22.5k; Total: 54.5k.}
    \label{fig:data-mixture-pie}
\end{figure}

\section{Evaluation}
As preprocessing for the evaluation, we removed all quotation marks from the source, reference, and candidate texts to prevent scores from being overly influenced by special character variations, e.g., single vs. double quotation marks.

Assessing \textsc{LuxMT}'s performance on the Luci benchmark, we find substantial improvements over the \textsc{Gemma 3} baseline (see Table~\ref{tab:performance} and Appendix~\ref{tab:full_deltas} for further metrics).
Interestingly, the scores also indicate an improvement of \textsc{LuxMT}'s performance in translating LB$\rightarrow$DE even though we did not fine-tune the model on any DE data. This finding is in line with literature demonstrating LLMs' cross-lingual learning capabilities \cite{philippy2023towards}.

\definecolor{virtuallycertain}{RGB}{20,180,120}   
\definecolor{verylikely}{RGB}{120,230,120}        
\definecolor{likely}{RGB}{200,255,120}             
\definecolor{aboutaslikely}{RGB}{255,255,120}    
\definecolor{unlikely}{RGB}{255,200,120}         
\definecolor{veryunlikely}{RGB}{255,140,120}      
\definecolor{exceptional}{RGB}{255,0,120}          
\newcommand{\accVirtuallyCertain}[1]{\cellcolor{virtuallycertain}#1}
\newcommand{\accVeryLikely}[1]{\cellcolor{verylikely}#1}
\newcommand{\accLikely}[1]{\cellcolor{likely}#1}
\newcommand{\accAboutAsLikely}[1]{\cellcolor{aboutaslikely}#1}
\newcommand{\accUnlikely}[1]{\cellcolor{unlikely}#1}
\newcommand{\accExceptional}[1]{\cellcolor{exceptional}\textcolor{white}{#1}}

\newcommand{\capBG}[3]{%
  {\setlength{\fboxsep}{0.4pt}%
   \colorbox{#1}{\strut\textcolor{#2}{#3}}}%
}

\newcommand{\capVirtuallyCertain}[1]{\capBG{virtuallycertain}{black}{#1}}
\newcommand{\capVeryLikely}[1]{\capBG{verylikely}{black}{#1}}
\newcommand{\capLikely}[1]{\capBG{likely}{black}{#1}}
\newcommand{\capAboutAsLikely}[1]{\capBG{aboutaslikely}{black}{#1}}
\newcommand{\capUnlikely}[1]{\capBG{unlikely}{black}{#1}}
\newcommand{\capVeryUnlikely}[1]{\capBG{veryunlikely}{black}{#1}}
\newcommand{\capExceptional}[1]{\capBG{exceptional}{black}{#1}}

\begin{table}[t]
\small
\centering
\begin{tabular}{lcccc}
\toprule
& \textsc{LE$\uparrow$}
& \textsc{BS$\uparrow$}
& \textsc{B-20$\uparrow$}
& \textsc{xC\textsuperscript{XL}$\uparrow$} \\
\midrule
{$\Delta$ LB$\rightarrow$FR}
& 0.7*
& \accAboutAsLikely{0.2*}
& \accLikely{1.3*}
& \accVeryLikely{1.9*} \\

\rowcolor{gray!20}
{$\Delta$ LB$\rightarrow$EN}
& 0.8*
& \accLikely{0.8*}
& \accLikely{0.9*}
& \accLikely{1.2*} \\

{$\Delta$ LB$\rightarrow$DE}
& 0.2
& \accAboutAsLikely{0.2*}
& \accLikely{1.4*}
& \accLikely{0.6*} \\

\bottomrule
\end{tabular}
\caption{System score deltas ($\Delta$) on the Luci benchmark. Asterisk (*) signifies statistical significance ($p < 0.05$) using paired bootstrap sampling with 95\% confidence interval. $\Delta = \textsc{LuxMT} - \textsc{Gemma 3}$ \\
Human accuracy estimates: \\
\capVirtuallyCertain{> 99\%};
\capVeryLikely{> 90\%};
\capLikely{> 66\%};
\capAboutAsLikely{33\%--66\%};
\capUnlikely{< 33\%};
\capVeryUnlikely{< 10\%};
\capExceptional{< 1\%}
}
\label{tab:performance}
\end{table}

\section{LuxEmbedder as QE Metric}
We compute Spearman's $\rho$ and Kendall's $\tau$ correlations over base models' system scores to assess the agreement between \textsc{LuxEmbedder} and the other automatic MT metrics mentioned above. We do not include \textsc{LuxMT} since we used \textsc{LuxEmbedder} to filter the training data, which would skew \textsc{LuxMT}'s LuxEmbedder score. We check the correlations for each language pair (LB$\rightarrow$FR, LB$\rightarrow$EN, LB$\rightarrow$DE). The results indicate that \textsc{LuxEmbedder} has the highest correlation with \textsc{chrF2}, followed by the other two surface overlap metrics (\textsc{BLEU}, \textsc{TER}) and \textsc{BERTScore} (see Appendix~\ref{sec:correlations}).
This finding is rather surprising, seeing as \textsc{LuxEmbedder} is more closely related to \textsc{BERTScore} than any surface overlap metric. The results suggest that \textsc{LuxEmbedder} should be used cautiously, as metrics trained on human preference data outperform surface-overlap metrics. Some researchers discourage using surface-overlap metrics for inter-system comparisons \cite{kocmi2024navigating}.

\section{Conclusion}
We introduced \textsc{LuxMT}, a machine translation system based on \textsc{Gemma 3}, specifically fine-tuned to translate LB to FR and EN. In addition, we constructed a novel multilingual human-translated benchmark for LB$\rightarrow$FR, LB$\rightarrow$EN, and LB$\rightarrow$DE.

Using \textsc{LuxEmbedder}, Luxembourgish sentence embeddings, we filtered training data and evaluated systems across a broad spectrum of automatic metrics. Our results demonstrate substantial performance gains not only for LB$\rightarrow$FR and LB$\rightarrow$EN, but also for LB$\rightarrow$DE even though the model was not fine-tuned on that language pair.

We also explored the use of \textsc{LuxEmbedder} as a QE metric for LB$\rightarrow$target-language and found that it correlates with reference-based metrics. However, we advise interpreting these scores with caution and highlight the need for further research to rigorously assess \textsc{LuxEmbedder}'s suitability for this purpose.

\section{Limitations}
\subsection{Benchmark}
\begin{itemize}
    \item \textbf{Domain-specificity}: The Luci benchmark is entirely based on a tourist magazine and thus limited in its domain-scope.
    \item \textbf{Size}: The benchmark is currently limited to a rather small size of 500 segment pairs per language pair which may impact the results' reliability. For example, \citet{kocmi2024navigating} find high variance in metric deltas under 500 segment pairs. In the future, we hope to expand the benchmark.
    \item \textbf{Source language}: It is not clear what the source language of Luci is. It is possible that the magazine was translated from FR, DE, or EN into LB. As a result, the LB text could contain translation artifacts, i.e., translationese \cite{gellerstam1986translationese}.
\end{itemize}

\subsection{Evaluation Metrics}
\begin{itemize}
    \item \textbf{Metric ensemble}: While we used a diverse range of evaluation metrics covering a broad spectrum of architectures and scoring system, we were unable to use \textsc{MetricX} \cite{juraska2025metricx}, a SOTA automatic metric, due to computation limitations. Furthermore, human evalutions remain the gold-standard and automatic metrics should be interpreted with caution.
    \item \textbf{LuxEmbedder as QE metric}: Using \textsc{LuxEmbedder} as a QE metric requires further research. While we found strong correlations between \textsc{LuxEmbedder} and other automatic MT metrics, we conducted our experiments on a small scale. We recommend using the metric with caution, as the results should be verified with human preference data. 

In the future, we will release linguistically-motivated test suites for a fine-grained evaluation of LB MT to analyse what linguistic phenomena the model learned through fine-tuning. We also plan to fine-tune the new \textsc{TranslateGemma} model, explore fine-tuning European models, and try filtering the data with estimated translation difficulty scores \cite{finkelstein2026translategemma, proietti2025estimating, kocmi2025command}.
\end{itemize}

\section*{Acknowledgments}
This research is part of the Lux-ASR 2 project\footnote{\url{https://infolux.uni.lu/lux-asr/}}, supported by the Chambre des Députés.

I would also like to thank my supervisor, Prof. Dr. Peter Gilles, for his guidance during this PhD project.

\bibliography{custom}

@inproceedings{philippy2025luxembedder,
  title={LuxEmbedder: A cross-lingual approach to enhanced Luxembourgish sentence embeddings},
  author={Philippy, Fred and Guo, Siwen and Klein, Jacques and Bissyande, Tegawende},
  booktitle={Proceedings of the 31st International Conference on Computational Linguistics},
  pages={11369--11379},
  year={2025}
}

@article{team2025gemma,
  title={Gemma 3 technical report},
  author={Team, Gemma and Kamath, Aishwarya and Ferret, Johan and Pathak, Shreya and Vieillard, Nino and Merhej, Ramona and Perrin, Sarah and Matejovicova, Tatiana and Ram{\'e}, Alexandre and Rivi{\`e}re, Morgane and others},
  journal={arXiv preprint arXiv:2503.19786},
  year={2025}
}

@article{moghe2025machine,
  title={Machine translation meta evaluation through translation accuracy challenge sets},
  author={Moghe, Nikita and Fazla, Arnisa and Amrhein, Chantal and Kocmi, Tom and Steedman, Mark and Birch, Alexandra and Sennrich, Rico and Guillou, Liane},
  journal={Computational Linguistics},
  volume={51},
  number={1},
  pages={73--137},
  year={2025},
  publisher={MIT Press 255 Main Street, 9th Floor, Cambridge, Massachusetts 02142, USA~…}
}

@article{kocmi2024navigating,
  title={Navigating the metrics maze: Reconciling score magnitudes and accuracies},
  author={Kocmi, Tom and Zouhar, Vil{\'e}m and Federmann, Christian and Post, Matt},
  journal={arXiv preprint arXiv:2401.06760},
  year={2024}
}

@inproceedings{papineni2002bleu,
  title={Bleu: a method for automatic evaluation of machine translation},
  author={Papineni, Kishore and Roukos, Salim and Ward, Todd and Zhu, Wei-Jing},
  booktitle={Proceedings of the 40th annual meeting of the Association for Computational Linguistics},
  pages={311--318},
  year={2002}
}

@inproceedings{popovic2015chrf,
  title={chrF: character n-gram F-score for automatic MT evaluation},
  author={Popovi{\'c}, Maja},
  booktitle={Proceedings of the tenth workshop on statistical machine translation},
  pages={392--395},
  year={2015}
}

@inproceedings{pu2021learning,
  title = {Learning compact metrics for MT},
  author = {Pu, Amy and Chung, Hyung Won and Parikh, Ankur P and Gehrmann, Sebastian and Sellam, Thibault},
  booktitle = {Proceedings of EMNLP},
  year = {2021}
}

@article{guerreiro2024xcomet,
  title={xcomet: Transparent machine translation evaluation through fine-grained error detection},
  author={Guerreiro, Nuno M and Rei, Ricardo and Stigt, Daan van and Coheur, Luisa and Colombo, Pierre and Martins, Andr{\'e} FT},
  journal={Transactions of the Association for Computational Linguistics},
  volume={12},
  pages={979--995},
  year={2024},
  publisher={MIT Press 255 Main Street, 9th Floor, Cambridge, Massachusetts 02142, USA~…}
}

@article{zhang2019bertscore,
  title={Bertscore: Evaluating text generation with bert},
  author={Zhang, Tianyi and Kishore, Varsha and Wu, Felix and Weinberger, Kilian Q and Artzi, Yoav},
  journal={arXiv preprint arXiv:1904.09675},
  year={2019}
}

@inproceedings{snover2006study,
  title={A study of translation edit rate with targeted human annotation},
  author={Snover, Matthew and Dorr, Bonnie and Schwartz, Richard and Micciulla, Linnea and Makhoul, John},
  booktitle={Proceedings of the 7th Conference of the Association for Machine Translation in the Americas: Technical Papers},
  pages={223--231},
  year={2006}
}

@article{costa2022no,
  title={No language left behind: Scaling human-centered machine translation},
  author={Costa-Juss{\`a}, Marta R and Cross, James and {\c{C}}elebi, Onur and Elbayad, Maha and Heafield, Kenneth and Heffernan, Kevin and Kalbassi, Elahe and Lam, Janice and Licht, Daniel and Maillard, Jean and others},
  journal={arXiv preprint arXiv:2207.04672},
  year={2022}
}

@inproceedings{popovic2016chrf,
  title={chrF deconstructed: beta parameters and n-gram weights},
  author={Popovi{\'c}, Maja},
  booktitle={Proceedings of the First Conference on Machine Translation: Volume 2, Shared Task Papers},
  pages={499--504},
  year={2016}
}

@article{xu2024benchmark,
  title={Benchmark data contamination of large language models: A survey},
  author={Xu, Cheng and Guan, Shuhao and Greene, Derek and Kechadi, M and others},
  journal={arXiv preprint arXiv:2406.04244},
  year={2024}
}

@article{abdin2024phi,
  title={Phi-4 technical report},
  author={Abdin, Marah and Aneja, Jyoti and Behl, Harkirat and Bubeck, S{\'e}bastien and Eldan, Ronen and Gunasekar, Suriya and Harrison, Michael and Hewett, Russell J and Javaheripi, Mojan and Kauffmann, Piero and others},
  journal={arXiv preprint arXiv:2412.08905},
  year={2024}
}

@article{dang2024aya,
  title={Aya expanse: Combining research breakthroughs for a new multilingual frontier},
  author={Dang, John and Singh, Shivalika and D'souza, Daniel and Ahmadian, Arash and Salamanca, Alejandro and Smith, Madeline and Peppin, Aidan and Hong, Sungjin and Govindassamy, Manoj and Zhao, Terrence and others},
  journal={arXiv preprint arXiv:2412.04261},
  year={2024}
}

@article{grattafiori2024llama,
  title={The llama 3 herd of models},
  author={Grattafiori, Aaron and Dubey, Abhimanyu and Jauhri, Abhinav and Pandey, Abhinav and Kadian, Abhishek and Al-Dahle, Ahmad and Letman, Aiesha and Mathur, Akhil and Schelten, Alan and Vaughan, Alex and others},
  journal={arXiv preprint arXiv:2407.21783},
  year={2024}
}

@inproceedings{juraska2025metricx,
  title={Metricx-25 and gemspaneval: Google translate submissions to the wmt25 evaluation shared task},
  author={Juraska, Juraj and Domhan, Tobias and Finkelstein, Mara and Nakagawa, Tetsuji and Kovacs, Geza and Deutsch, Daniel and Wang, Pidong and Freitag, Markus},
  booktitle={Proceedings of the Tenth Conference on Machine Translation},
  pages={957--968},
  year={2025}
}

@article{gellerstam1986translationese,
  title={Translationese in Swedish novels translated from English},
  author={Gellerstam, Martin},
  journal={Translation studies in Scandinavia},
  volume={1},
  pages={88--95},
  year={1986}
}

@article{philippy2023towards,
  title={Towards a common understanding of contributing factors for cross-lingual transfer in multilingual language models: A review},
  author={Philippy, Fred and Guo, Siwen and Haddadan, Shohreh},
  journal={arXiv preprint arXiv:2305.16768},
  year={2023}
}

@inproceedings{koehn2004statistical,
  title={Statistical significance tests for machine translation evaluation},
  author={Koehn, Philipp},
  booktitle={Proceedings of the 2004 conference on empirical methods in natural language processing},
  pages={388--395},
  year={2004}
}

@article{finkelstein2026translategemma,
  title={TranslateGemma Technical Report},
  author={Finkelstein, Mara and Caswell, Isaac and Domhan, Tobias and Peter, Jan-Thorsten and Juraska, Juraj and Riley, Parker and Deutsch, Daniel and Dilanni, Cole and Cherry, Colin and Briakou, Eleftheria and others},
  journal={arXiv preprint arXiv:2601.09012},
  year={2026}
}

@article{proietti2025estimating,
  title={Estimating machine translation difficulty},
  author={Proietti, Lorenzo and Perrella, Stefano and Zouhar, Vil{\'e}m and Navigli, Roberto and Kocmi, Tom},
  journal={Preprint},
  year={2025}
}

@inproceedings{kocmi2025command,
  title={Command-A-Translate: Raising the Bar of Machine Translation with Difficulty Filtering},
  author={Kocmi, Tom and Arkhangorodsky, Arkady and Berard, Alexandre and Blunsom, Phil and Cahyawijaya, Samuel and Dehaze, Th{\'e}o and Fadaee, Marzieh and Frosst, Nicholas and Galle, Matthias and Gomez, Aidan and others},
  booktitle={Proceedings of the Tenth Conference on Machine Translation},
  pages={789--799},
  year={2025}
}

\onecolumn

\appendix
\clearpage

\section{Full Results per Language Pair}
\label{sec:full-results}

\begin{longtable}{lccccccc}
\toprule
 &
\textsc{LE}$\uparrow$ &
\textsc{BS}$\uparrow$ &
\textsc{B-20}$\uparrow$ &
\textsc{xC\textsuperscript{XL}}$\uparrow$ &
\textsc{BL}$\uparrow$ &
\textsc{cF2}$\uparrow$ &
\textsc{TER}$\downarrow$ \\
\midrule
\multicolumn{8}{c}{LB$\rightarrow$FR} \\
\cmidrule(r){1-8} 
BL: \textsc{Gemma 3 27B IT Q8} &
\textbf{92.7} &
\textbf{95.2} &
\textbf{68.5} &
\textbf{85.1} &
\textbf{33.5} &
\textbf{61.2} &
\textbf{56.1} \\
\rowcolor{gray!20}
\textsc{Aya Expanse 32B Q8} &
87.7* &
94.0* &
57.7* &
72.8* &
26.3* &
54.9* &
64.3* \\
\textsc{Command R 35B Q8 (08-2024)} &
87.5* &
94.0* &
58.0* &
73.3* &
24.6* &
54.2* &
65.7* \\
\rowcolor{gray!20}
\textsc{Llama 3.1 70B IT Q5 K M} &
91.4* &
94.6* &
63.7* &
80.8* &
27.7* &
57.2* &
60.5* \\
\textsc{Llama 4 16x17B} &
92.4 &
94.7* &
64.1* &
80.3* &
29.8* &
57.9* &
59.7* \\
\rowcolor{gray!20}
\textsc{Mistral S 3.2 24B IT 2506 FP16} &
88.8* &
94.1* &
57.9* &
72.2* &
27.3* &
55.3* &
63.3* \\
\textsc{Phi 4 14B FP16} &
83.1* &
92.7* &
50.4* &
60.3* &
18.3* &
50.7* &
90.7* \\
\cmidrule(r){1-8} 
\multicolumn{8}{c}{LB$\rightarrow$EN} \\
\cmidrule(r){1-8} 
BL: \textsc{Gemma 3 27B IT Q8} &
\textbf{93.3} &
\textbf{88.5} &
\textbf{73.3} &
\textbf{88.1} &
\textbf{38.8} &
\textbf{64.2} &
\textbf{50.0} \\
\rowcolor{gray!20}
\textsc{Aya Expanse 32B Q8} &
89.4* &
85.0* &
68.3* &
81.4* &
31.9* &
58.5* &
57.8* \\
\textsc{Command R 35B Q8 (08-2024)} &
87.9* &
84.1* &
67.3* &
79.6* &
27.4* &
56.0* &
61.5* \\
\rowcolor{gray!20}
\textsc{Llama 3.1 70B IT Q5 K M} &
92.5* &
87.5* &
71.2* &
86.1* &
35.0* &
61.2* &
53.7* \\
\textsc{Llama 4 16x17B} &
92.5* &
87.2* &
70.8* &
85.5* &
35.4* &
61.2* &
53.8* \\
\rowcolor{gray!20}
\textsc{Mistral S 3.2 24B IT 2506 FP16} &
89.5* &
84.4* &
66.7* &
78.7* &
30.5* &
57.6* &
59.4* \\
\textsc{Phi 4 14B FP16} &
88.7* &
83.6* &
66.4* &
77.7* &
26.3* &
55.4* &
64.3* \\
\cmidrule(r){1-8} 
\multicolumn{8}{c}{LB$\rightarrow$DE} \\
\cmidrule(r){1-8} 
BL: \textsc{Gemma 3 27B IT Q8} &
\textbf{95.8} &
\textbf{97.2} &
\textbf{80.1} &
\textbf{95.8} &
\textbf{56.7} &
\textbf{77.3} &
\textbf{30.7} \\
\rowcolor{gray!20}
\textsc{Aya Expanse 32B Q8} &
93.0* &
96.1* &
74.9* &
92.4* &
46.0* &
70.5* &
40.6* \\
\textsc{Command R 35B Q8 (08-2024)} &
92.2* &
95.9* &
73.8* &
92.0* &
43.9* &
69.6* &
43.1* \\
\rowcolor{gray!20}
\textsc{Llama 3.1 70B IT Q5 K M}&
95.2* &
96.8* &
78.2* &
95.1* &
51.5* &
74.8* &
34.8* \\
\textsc{Llama 4 16x17B} &
95.6 &
96.8* &
77.7* &
94.2* &
52.5* &
75.2* &
34.9* \\
\rowcolor{gray!20}
\textsc{Mistral S 3.2 24B IT 2506 FP16} &
92.8* &
96.2* &
72.8* &
90.4* &
48.4* &
72.3* &
38.5* \\
\textsc{Phi 4 14B FP16} &
85.0* &
94.4* &
66.7* &
82.5* &
28.9* &
63.8* &
85.9* \\
\bottomrule
\caption{BL = baseline. Automatic evaluation results using \textsc{LuxEmbedder} (LE), \textsc{BERTScore} (BS), \textsc{BLEURT-20} (B-20), \textsc{xCOMET\textsuperscript{XL}} (xC\textsuperscript{XL}), \textsc{BLEU} (BL), \textsc{chrF2} (cF2), and \textsc{Translation Edit Rate} (TER) on the Luci benchmark. The best scores are in bold. Asterisks denote statistically significant differences from the \textsc{Gemma 3} (BS) baseline ($p < 0.05$).}
\end{longtable}

\clearpage

\FloatBarrier

\section{Temperature Exploration}
\label{sec:appendix_temp}

\begin{longtable}{lccccc}
\toprule
Temp &
\textsc{BS}$\uparrow$ &
\textsc{B-20}$\uparrow$ &
\textsc{BL}$\uparrow$ &
\textsc{cF2}$\uparrow$ &
\textsc{TER}$\downarrow$ \\
\midrule
0.1 & 94.3 & \textbf{78.7} & 44.7 & \textbf{68.8} & \textbf{43.6} \\
\rowcolor{gray!20}
0.2 & 94.3 & 78.6 & 44.6 & \textbf{68.8} & 44.2 \\
0.3 & \textbf{94.4} & 78.5 & 44.6 & 68.6 & 43.7 \\
\rowcolor{gray!20}
0.4 & 94.2 & 78.3 & 44.0 & 68.2 & 44.4 \\
0.5 & 94.3 & 78.5 & 43.8 & 68.3 & 44.8 \\
\rowcolor{gray!20}
0.6 & 94.3 & 78.5 & 43.9 & 68.6 & 44.4 \\
0.7 & 94.3 & 78.6 & 44.3 & 68.5 & 44.2 \\
\rowcolor{gray!20}
0.8 & 94.3 & \textbf{78.7} & \textbf{44.8} & \textbf{68.8} & 44.1 \\
0.9 & 94.3 & 78.6 & 44.7 & \textbf{68.8} & 44.0 \\
\rowcolor{gray!20}
1.0 & \textbf{94.4} & \textbf{78.7} & 44.6 & 68.7 & 44.0 \\
\bottomrule
\caption{Automatic evaluation scores across different temperature settings. The best scores are in bold.}
\label{tab:temperature_sweep}
\end{longtable}

\FloatBarrier

\section{Automatic Evaluation Results across Various \textsc{LuxEmbedder} Filter Thresholds}
\label{sec:automatic_eval_luxembedder}

\begin{longtable}{lccccc}
\toprule
System & \textsc{BS}$\uparrow$ & \textsc{B-20}$\uparrow$ & \textsc{BL}$\uparrow$ & \textsc{cF2}$\uparrow$ & \textsc{TER}$\downarrow$ \\
\midrule
\textsc{Gemma 3} & 93.8 & 77.2 & 42.5 & 67.8 & 46.3 \\
\rowcolor{gray!20}
\textsc{LE} .90      & 93.6 & 77.6 & 40.4 & 65.6 & 48.2 \\
\textsc{LE} .95      & 94.0 & 78.1 & 42.0 & 66.5 & 46.3 \\
\rowcolor{gray!20}
\textsc{LE} .99      & \textbf{94.4} & \textbf{78.8} & \textbf{44.7} & \textbf{69.1} & \textbf{43.6} \\
\bottomrule
\caption{Automatic evaluation results assessing various \textsc{LuxEmbedder} (LE) filter thresholds, using \textsc{BERTScore} (BS), \textsc{BLEURT-20} (B-20), \textsc{BLEU} (BL), \textsc{chrF2} (cF2), and \textsc{Translation Edit Rate} (TER). For all metrics, higher scores are better ($\uparrow$), except for \textsc{TER} where lower is better ($\downarrow$). Results are based on average scores for LB$\rightarrow$FR ($N=400$) and LB$\rightarrow$EN ($N=233$) on Luci segment pairs.}
\end{longtable}

\FloatBarrier

\section{Automatic Evaluation Results Assessing Fine-Tuning Schedule}
\label{sec:appendix_schedule}

\begin{longtable}{lccccc}
\toprule
Epochs & \textsc{BS}$\uparrow$ & \textsc{B-20}$\uparrow$ & \textsc{BL}$\uparrow$ & \textsc{cF2}$\uparrow$ & \textsc{TER}$\downarrow$ \\
\midrule
1     & \textbf{94.4} & \textbf{78.8} & \textbf{44.7} & \textbf{69.1} & \textbf{43.6} \\
\rowcolor{gray!20}
2     & \textbf{94.4} & 78.5 & 44.1 & 68.4 & 44.4 \\
3      & 94.2 & 78.4 & 43.6 & 68.1 & 44.7 \\
\bottomrule
\caption{Automatic evaluation results assessing fine-tuning \textsc{Gemma 3} for 1-3 epochs. Results are based on average system scores for LB$\rightarrow$FR ($N=400$) and LB$\rightarrow$EN ($N=233$) on Luci segment pairs.}
\end{longtable}

\FloatBarrier
\clearpage

\section{System Score Deltas on the Luci Benchmark}

\begin{longtable}{lccccccc}
\toprule
& \textsc{LE}$\uparrow$
& \textsc{BS}$\uparrow$
& \textsc{B-20}$\uparrow$
& \textsc{xC\textsuperscript{XL}}$\uparrow$
& \textsc{BL}$\uparrow$
& \textsc{cF2}$\uparrow$
& \textsc{TER}$\downarrow$ \\
\midrule
{$\Delta$ LB$\rightarrow$FR}
& 0.7*
& \accAboutAsLikely{0.2*}
& \accLikely{1.3*}
& \accVeryLikely{1.9*}
& \accAboutAsLikely{0.6}
& \accAboutAsLikely{0.4}
& -1.6* \\
\rowcolor{gray!20}
{$\Delta$ LB$\rightarrow$EN}
& 0.8*
& \accLikely{0.8*}
& \accLikely{0.9*}
& \accLikely{1.2*}
& \accLikely{1.2*}
& \accLikely{1.3*}
& -2.3* \\

{$\Delta$ LB$\rightarrow$DE}
& 0.2
& \accAboutAsLikely{0.2*}
& \accLikely{1.4*}
& \accLikely{0.6*}
& \accAboutAsLikely{0.9*}
& \accAboutAsLikely{0.3}
& -0.7 \\

\bottomrule
\caption{System score deltas ($\Delta$) on the Luci benchmark. Asterisk (*) signifies statistical significance ($p < 0.05$) using paired bootstrap sampling with 95\% confidence interval. $\Delta = \textsc{LuxMT} - \textsc{Gemma 3}$ \\
Human accuracy estimates: \\
\capVirtuallyCertain{> 99\%};
\capVeryLikely{> 90\%};
\capLikely{> 66\%};
\capAboutAsLikely{33\%--66\%};
\capUnlikely{< 33\%};
\capVeryUnlikely{< 10\%};
\capExceptional{< 1\%}
}
\label{tab:full_deltas}
\end{longtable}

\FloatBarrier

\section{Automatic Metric Correlations with LuxEmbedder}
\label{sec:correlations}

\begin{longtable}{lcc}
\toprule
&
Spearman's $\rho$ &
Kendall's $\tau$ \\
\midrule
\multicolumn{3}{c}{LB$\rightarrow$FR} \\
\cmidrule(r){1-3}
\textsc{BERTScore}
& 0.9910*
& 0.9759* \\

\rowcolor{gray!20} 
\textsc{BLEURT-20}
& 0.8929*
& 0.8095* \\

\textsc{xCOMET\textsuperscript{XL}}:
& 0.8214*
& 0.6190* \\

\rowcolor{gray!20} 
\textsc{BLEU}
& \textbf{1.0000*}
& \textbf{1.0000*} \\

\textsc{chrF2}
& \textbf{1.0000*}
& \textbf{1.0000*} \\

\rowcolor{gray!20} 
\textsc{TER}
& \textbf{1.0000*}
& \textbf{1.0000*} \\

\cmidrule(r){1-3}
\multicolumn{3}{c}{LB$\rightarrow$EN} \\
\cmidrule(r){1-3}

\textsc{BERTScore}
& 0.9190*
& 0.7807* \\

\rowcolor{gray!20} 
\textsc{BLEURT-20}
& 0.8108*
& 0.6831* \\

\textsc{xCOMET\textsuperscript{XL}}
& 0.8108*
& 0.6831* \\

\rowcolor{gray!20} 
\textsc{BLEU}
& 0.9190*
& 0.7807* \\

\textsc{chrF2}
& \textbf{0.9273*}
& \textbf{0.8000*} \\

\rowcolor{gray!20} 
\textsc{TER}
& 0.9190*
& 0.7807* \\

\cmidrule(r){1-3}
\multicolumn{3}{c}{LB$\rightarrow$DE} \\
\cmidrule(r){1-3}

\textsc{BERTScore}
& 0.9550*
& 0.8783* \\

\rowcolor{gray!20} 
\textsc{BLEURT-20}
& 0.9286*
& 0.8095* \\

\textsc{xCOMET\textsuperscript{XL}}
& 0.9286*
& 0.8095* \\

\rowcolor{gray!20} 
\textsc{BLEU}
& \textbf{0.9643*}
& \textbf{0.9048*} \\

\textsc{chrF2}
& \textbf{0.9643*}
& \textbf{0.9048*} \\

\rowcolor{gray!20} 
\textsc{TER}
& 0.9286*
& 0.8095* \\

\bottomrule
\caption{Correlation of \textsc{LuxEmbedder} and other automatic metrics on system-level performance scores of \textsc{Gemma 3}, \textsc{Aya Expanse}, \textsc{Command R}, \textsc{Llama 3.1}, \textsc{Llama 4}, \textsc{Mistral S}, and \textsc{Phi 4} on the Luci benchmark ($N = 500$ for each language pair). Asterisk (*) denotes statistical significance ($p < 0.05$) and the highest correlations are marked in bold.}
\end{longtable}

\end{document}